\documentclass[a4paper,11pt]{article}
\usepackage{gensymb}
\usepackage{fourier}
\usepackage{color}
 \usepackage{graphicx}
\usepackage{url}
\usepackage[affil-it]{authblk}
\usepackage{amsmath}
\usepackage{wrapfig}
\usepackage{lipsum,booktabs}
\usepackage{enumitem}
\usepackage[T1]{fontenc}
\usepackage{times}
\usepackage{multirow}
\usepackage[justification=centering]{caption}
\begin{document}

\title{Using Filter Banks in Convolutional Neural Networks for Texture Classification}

\author{V. Andrearczyk \& Paul F. Whelan}
\affil{Vision Systems Group, School of Electronic Engineering, Dublin City University, Glasnevin, Dublin 9, Ireland}
\date{}
\maketitle
\thispagestyle{empty}

\begin{abstract}
Deep learning has established many new state of the art solutions in the last decade in areas such as object, scene and speech recognition.
In particular Convolutional Neural Network (CNN) is a category of deep learning which obtains excellent results in object detection and recognition tasks.
Its architecture is indeed well suited to object analysis by learning and classifying complex (deep) features that represent parts of an object or the object itself.
However, some of its features are very similar to texture analysis methods. CNN layers can be thought of as filter banks of complexity increasing with the depth.
Filter banks are powerful tools to extract texture features and have been widely used in texture analysis.
In this paper we develop a simple network architecture named Texture CNN (T-CNN) which explores this observation.
It is built on the idea that the overall shape information extracted by the fully connected layers of a classic CNN is of minor importance in texture analysis.
Therefore, we pool an energy measure from the last convolution layer which we connect to a fully connected layer.
We show that our approach can improve the performance of a network while greatly reducing the memory usage and computation.

\end{abstract}

\let\thefootnote\relax\footnote{\em Article published in Pattern Recognition Letters. DOI: 10.1016/j.patrec.2016.08.016\hspace{1.5in}16/08/2016}
\section{Introduction}
Texture, together with color is a key component in the analysis of images.
Although there is no generally accepted definition for texture in computer vision, it is established that a texture image or region obeys some statistical properties and exhibits repeated structures \cite{haralick1973textural}, \cite{leung2001representing}.
Texture analysis approaches such as statistical, structural, model-based or transform-based exploit these observations to classify, segment or synthesize textured images \cite{bharati2004image}.
\\Pioneered, among others, by Fukushima \cite{fukushima1980neocognitron}, Lecun \cite{lecun1998gradient} and Hinton \cite{krizhevsky2012imagenet}, Convolutional Neural Networks (CNNs) have been generalized since the breakthrough in the 2012
ImageNet Large Scale Visual Recognition Challenge \cite{russakovsky2014imagenet} of Krizhevsky \cite{krizhevsky2012imagenet}, and improved the state of the art of many machine vision tasks.
\\The complexity of the features trained by CNNs increases with the depth of the network. Therefore the last convolution layer extracts complex features
which respond to objects such as a nose, a face or a human body.
The fully connected layers use the response to these features to obtain information about the overall shape of the image and calculate a probability distribution over the different classes in the last fully connected layer.
This design is suitable for exploring the arrangement of less complex features from the previous layers and their sparse spatial response for an object recognition scheme.
This overall shape analysis and the sparsity and complexity of features is less adequate in texture analysis as we mainly seek repeated patterns of lower complexity.
However, the dense orderless extraction of features by the intermediate layers using weight sharing is of high interest in texture analysis.
The first layer extracts edge-like features and can be thought of as a filter bank approach such as Gabor filters \cite{fogel1989gabor} or Maximum Response filters \cite{caenen2004maximum},\cite{varma2005statistical}, widely used in texture analysis.
Intermediate convolution and pooling layers are alike filter banks extracting features of increasing complexity.
It is possible that a classic CNN architecture could learn how to explore the properties of texture images most efficiently without modifying its structure
when trained on a very large texture dataset (such as ImageNet). This assumption can not be verified as, to the best of our knowledge, such a database does not exist.

We build upon \cite{cimpoi2015deep} to create a CNN designed to learn and classify texture images.
We are interested in creating a CNN that fully incorporates the learning of texture descriptors and their classification as opposed to the Fisher Vector CNN (FV-CNN). Therefore, we pool simple energy descriptors from convolution layers
that are used inside the CNN and allow forward and backward propagation to learn texture features.
While Cimpoi \cite{cimpoi2015deep} insists on the domain transferability of CNNs by demonstrating that a network pre-trained on ImageNet can be used to accurately classify texture images, we try to analyze
a texture specific domain and the result of training a network only on texture images. Therefore we evaluate trainings from scratch as well as networks pre-trained on both texture and object datasets.

We demonstrate that simple networks with reduced number of neurons and weights are able to achieve similar or better results on texture recognition datasets.
One of the major trends in the community of deep neural networks is to use more and more complex networks, using the increasing power and memory of computers and GPUs to train very deep and computationally expensive networks.
However, the interest of CNNs is not limited to powerful desktop computers and designing efficient networks while restraining their size is important for mobile and embedded computing as mentioned in \cite{szegedy2014going}.
In particular, we are interested in training networks from scratch on datasets of various sizes which is more computationally expensive than using pre-trained CNNs.
In consequence, this research is not focused on competing with the state of the art in texture recognition but rather on designing a simple architecture which explores the ideas introduced in this section
 as well as gaining insight on how CNNs learn texture feature.
\\ To summarize, the main contributions of this paper are as follows: (a) We create a simple CNN with reduced complexity which extracts, learns and classifies texture features;
(b) We evaluate the performance of networks from scratch and pre-trained ones, as well as the domain transferability of the latter;
(c) We experiment various depth networks when applied to texture and object datasets;
(d) We combine texture and shape anlysis within a new network architecture.
\\The rest of the paper is organized as follows. In section 2, we describe the related work.
In section 3, we introduce the architecture of our proposed texture network and its combination with a classic CNN.
In section 4, we describe our experimental protocol and discuss our results.

\section{Related work}
Basic CNN architectures have been applied to texture recognition such as \cite{tivive2006texture}, in which a simple four layers network was used in the early stage of deep learning  to classify the Brodatz database. 
More recently, Hafemann \cite{hafemann2014forest} applied CNN to a forest species classification, similar to a texture classification problem. While more complex and more accurate than \cite{tivive2006texture},
this approach still does not take the characteristics of texture images (statistical properties and repeated patterns) into consideration as it is a simple application of a standard CNN to a texture dataset.
\\Cimpoi \cite{cimpoi2015deep} demonstrated the relevance of densely extracting texture descriptors from a CNN with the FV-CNN. They obtain impressive results on both
texture recognition and texture recognition in clutter datasets. This approach is well suited to region recognition as it requires computing the convolution layer output once and pooling regions with
FV separately.
However, the CNN part of the FV-CNN does not learn from the texture dataset. A pre-trained network extracts the outputs of the convolution layers without being finetuned on the texture dataset.
The fully connected layers are replaced with FV encoding and SVM classification, also making this approach computationally expensive.

\section{Method description}
This section describes our Texture CNN (T-CNN) architecture and its combination with a classic deep neural network approach.

\subsection{Texture Convolutional Neural Network (T-CNN)}
\begin{figure}[!h]
\begin{center}
\includegraphics[width=1\textwidth]{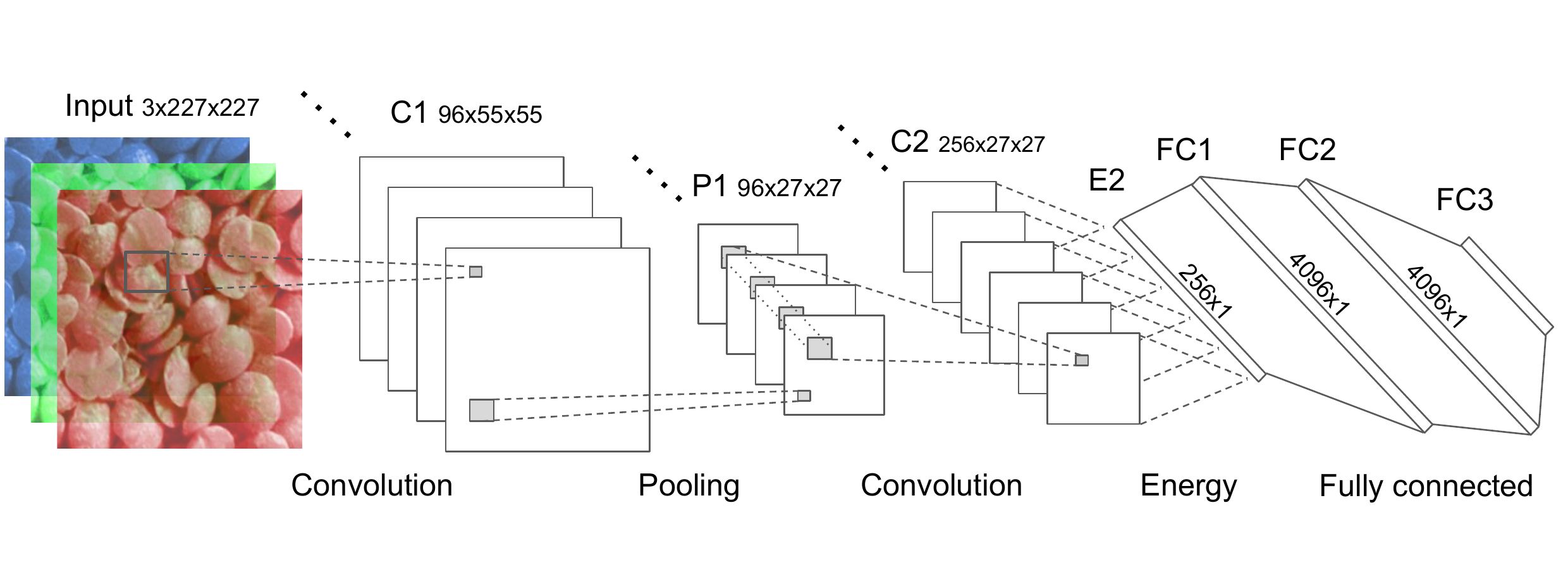}
\end{center}
\vspace{-20pt}
  \caption{T-CNN architecture using two convolution layers}\label{fig:TCNN_color_C2}
  \vspace{-10pt}
\end{figure}

CNNs naturally pool dense orderless features by the weight sharing of the convolution layers.
These layers can be compared to filter banks widely used in texture analysis which extract the response to a set of features.
While these features are pre-designed in fiter banks methods, the power of CNNs is to be able to learn appropriate features through forward and backward propagation.
These learned features can be combined within the network to classify new unknown images.
We develop a simple network architecture which explores this attribute. As explained in \cite{cimpoi2015deep}, the global spatial information is of minor importance
in texture analysis as opposed to the necessity of analyzing the global shape for an object recognition scheme.
Therefore, we want to pool dense orderless texture descriptors from the output of a convolution layer.
Our network is derived from AlexNet \cite{krizhevsky2012imagenet}. We develop a new energy layer as described below and use multiple configurations with varying numbers of convolution layers.
Each feature map of the last convolution layer is simply pooled by calculating the average
of its rectified linear activation output. This results in one single value per feature map, similar to an energy response to a filter bank where we use learned features of varying complexity instead of fixed filters.
A configuration with two convolution layers C1 and C2 is shown in Figure \ref{fig:TCNN_color_C2}.
The forward and backward propagation of the energy layer E2 are similar to an average pooling layer over the entire feature map, i.e. with a kernel size equal to the size of the feature map.
The vector output of the energy layer is simply connected to a fully connected layer FC1, followed by two other fully connected layers FC2 and FC3.
Similar to other networks, FC3 is of size equal to the number of classes and the probabilities of the classes are calculated with a Softmax layer. We use linear rectification, normalization and dropout in the same way as AlexNet.
We experiment with five configurations T-CNN-1 to T-CNN-5 with one to five convolution layers respectively. The energy is consistently pooled from the output of the last convolution layer.

It is important to notice that the number of parameters in our network is significantly lower than in a classic CNN such as AlexNet. Not only are convolution layers removed (in T-CNN-1 to T-CNN-4),
but also the number of connections between the energy layer E2 and the fully connected layer FC3 is much smaller than the classic full connection of the last convolution layer.
The number of trainable parameters of the different networks used in this paper are indicated in Tables \ref{tab:scratch1}, \ref{tab:finetune1} and \ref{tab:combined}.
These are the total number of weights and biases of the networks applied to ImageNet (1000 classes).
Also, the size of the fully connected layers can be drastically reduced without deteriorating the performance as they encode a much smaller amount of neurons from E2.
Therefore, the number of parameters to train is reduced, as is the training and testing computational time as well as the required memory.

\subsection{Combining texture and classic CNNs}
\begin{figure}[!h]
\begin{center}
\includegraphics[width=1\textwidth]{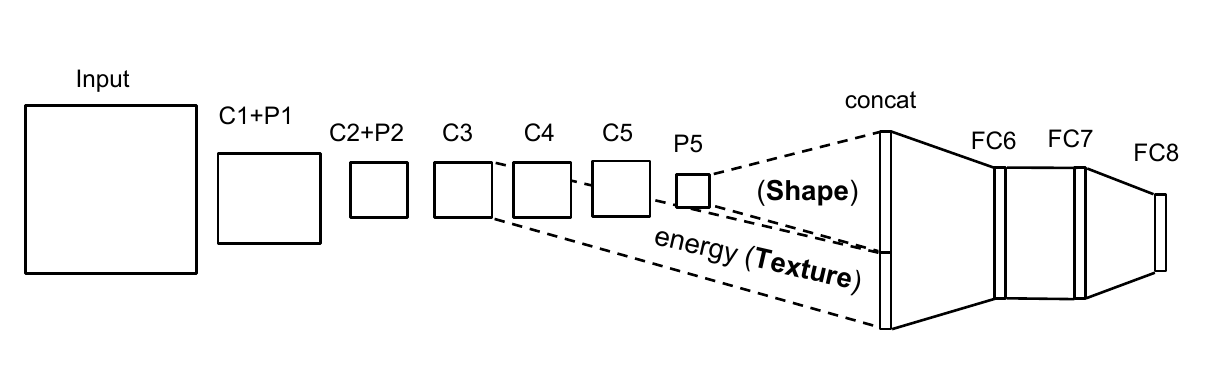}
\end{center}
\vspace{-20pt}
 \caption{Architecture of the Texture and Shape CNN (TS-CNN-3), integrating T-CNN-3 to a classic CNN.}\label{fig:TCNN_Original_C3}
  \vspace{-10pt}
\end{figure}

The design of the T-CNN enables a simple and efficient integration into a classic CNN architecture.
Figure \ref{fig:TCNN_Original_C3} illustrates this new architecture in which the energy layer of the T-CNN-3 is extracted within the classic CNN
and its output is concatenated to the flattened output of the last convolution layer.
In Caffe \cite{jia2014caffe}, we use existing layers to first flatten the outputs of the energy layer and of the last pooling layer P5 separately.
We then concatenate these flattened outputs using a concatenation layer and connect it to the fully connected layers without modifying the latter from the AlexNet layers.
This combination resembles the Multi-Stage features developed in \cite{sermanet2012convolutional}.
In this way, the network analyzes the texture and the overall shape of the image, likewise \cite{cimpoi2015deep}
in which the FV-CNN for texture analysis is combined to the Fully-Connected CNN (FC-CNN) for the overall shape analysis. However, our method incorporates the two approaches within the same network which learns and combines texture and shape features.
It is important to note that the texture and shape analysis share the same previous and following layers (C1, C2, C3 and FC6, FC7, FC8),
thus keeping the computation time and memory consumption close to the original AlexNet as shown in Table \ref{tab:combined} .

\section{Results and analysis}
\subsection{Details of the network}
We use Caffe \cite{jia2014caffe} to implement our network\footnote{An implementation to train (scratch and fine-tune) and test our T-CNN-3 on kth-tips-2b is provided here:\\https://github.com/v-andrearczyk/caffe-TCNN}, where our architecture is derived from AlexNet.
We keep the same number of feature maps, kernel sizes etc. for comparison.
However, it is possible to reduce the size of the fully connected layers of the T-CNN by a factor greater than two in average without loss of accuracy.
In general the base learning rate is 0.001 for networks that learn from scratch and 0.0001 for finetuning; the weight decay is 0.0005.
Yet, this is not a fixed rule and even though it is relatively stable, we must adapt the hyperparameters to the experiments (number of training samples, depth, finetuning or from scratch).
The results are also robust to small variations of the batch size but we also adapt the latter to the training sizes.
We use a batch size of 32 for the smallest training sets and up to 256 for ImageNet.
Finally, we keep the cropping of the input images to 227x227 for the sake of comparison for input images larger than 227x227. 

\subsection{Datasets}
We use a total of ten datasets; seven are texture datasets, the other three are object datasets as described below.

The \textit{ImageNet} 2012 dataset \cite{russakovsky2014imagenet} contains 1000 classes. The training set contains 1,281,167 images and the validation set (used for testing) 50,000 images (50 images/class) of size 256x256.
\\\textbf{Subsets of ImageNet:} 
We want to compare networks pre-trained on texture and object databases of the same size when it comes to finetuning these networks on another texture dataset.
Therefore, we create three subsets of ImageNet. For each subset, we select 28 classes from ImageNet. For each selected class, we keep the same training and testing images as ImageNet.
Therefore, each dataset contains 1400 test images and approximately 36000 images depending on the selected classes.
The full list of classes selected for each subset can be found in appendix A, B and C.

\textit{ImageNet-T} is a subset which retains texture classes such as ``stone wall'', ``tile roof'' and ``velvet''. Based on visual examination, 28 classes with high texture appearance are selected.

\textit{ImageNet-S1} is another subset with chosen object-like classes such as ``chihuahua'', ``ambulance'' and ``hammer''. Based on visual examination, classes of object images are selected.

Finally, in the last subset \textit{ImageNet-S2}, the 28 classes are chosen randomly.

\textit{kth-tips-2b} \cite{hayman2004significance} contains 11 classes of 432 texture images. Each class is made of four samples (108 images/sample).
Each sample is used once as a training set while the remaining three samples are used for testing. 
The results for kth-tips-2b are given as the mean classification and the standard deviation over the four splits.
The images are resized to 227x227. We do not resize the images to 256x256 followed by a cropping as the original images are smaller than 227x227. 

\textit{Kylberg} \cite{Kylberg2011c} is a texture database containing 28 classes of 160 images of size 576x576. We replicate the setup of \cite{kylberg2013evaluation}.
We randomly choose one of 12 available orientations for each image and split the images into four subimages which results in 28x160x4 = 17920 images of size 288x288 which we resize to 256x256.
We use a tenfolded cross-validation and report the average results. The cross-validation folds are created once and kept fixed throughout the experiments.

\textit{CUReT} \cite{dana1999reflectance} is a texture database which contains 61 classes.
We reproduce the setup of \cite{cimpoi2014describing} in which 92 images are used per class with 46 for training, 46 for testing.
We resize the 200x200 images to 227x227. We do not resize the images to 256x256 followed by a cropping as the original images are smaller than 227x227. 
The experiment is repeated 20 times and we report the results averaged over the 20 splits.

\textit{DTD} \cite{cimpoi2014describing} contains 47 classes of 120 images ``in the wild'' each.
The images are of various sizes and even though using multiple input sizes should be possible with our T-CNN, we resize all the images to 227x227 for comparison with AlexNet which requires fixed input images.
The dataset includes 10 available annotated splits with 40 training images, 40 validation images and 40 testing images for each class.
The results are then averaged over the 10 splits.

\textit{Macroscopic} \cite{paula2014forest} and \textit{Microscopic forest species databases} \cite{martins2013database} are also used to test our approach on larger texture images (respectively 3264x2448 and 1024x768).
We resize the images of both datasets to 640x640.
The experimental setup is reproduced from \cite{hafemann2014forest} except that we average the results over ten trials instead of three. 

\subsection{Results of Texture Convolutional Neural Network (T-CNN)}
\subsubsection{Networks from scratch and pre-trained}
Table \ref{tab:scratch1} shows the results of training the T-CNN from scratch using one to five convolution layers.
Table \ref{tab:finetune1} presents the classification rates using the networks pre-trained on the ImageNet database.
In both cases, our approach (T-CNN-3) performs almost always better than AlexNet on the texture datasets: Kylberg, CUReT, DTD, kth-tips-2b and ImageNet-T;
while containing nearly three times fewer trainable parameters (see Table \ref{tab:scratch1}).
The largest improvement is for the kth-tips-2b dataset on which our fine-tuned method outperforms AlexNet by 1.7\% (73.2\% and 71.5\% respectively).
The original AlexNet only slightly outperforms T-CNN-3 on the CUReT dataset from scratch (see Table \ref{tab:scratch1})
and on ImageNet-T and DTD when using the network pre-trained on ImageNet (see Table \ref{tab:finetune1}).
\\T-CNN-3 also performs best from scratch on the object-like subsets of ImageNet: ImageNet-S1 and ImageNet-S2.
We believe that our network can learn from the texture regions present in these images.
\\Our method does not compete with the state of the art as seen in Table \ref{tab:finetune1} because the latter use much more complex and deeper architectures. For this reason, the comparison to AlexNet is more meaningful.
One can notice that T-CNN-1 performs better on ImageNet-T which contains texture images (42.7\%) and ImageNet-S2 (42.1\%)
which is a random selection of classes than on ImageNet-S1 (34.9\%) which is a selection of classes containing object-like images.
This is due to the fact that the first convolution layer extracts extremely simple features (mainly edges) and acts like a Gabor filter which is not robust at classifying object datasets.
\\However, when using pre-trained networks, the original AlexNet expectedly performs better on non-texture images (ImageNet-S1 and ImageNet-S2) as it is designed
for object recognition and can learn more object-like features from the large database.
\\One could think that the reason we obtain good results is that small networks are well suited to small datasets
as the low number of parameters prevents overfitting while being sufficient to learn appropriate features.
Whereas very large networks are generally preferred to train large training sets using dropout to avoid overfitting.
However, using a shallow network based on AlexNet without using our energy layer results in a drop of accuracy as compared to our method.
We have tested a shortened version of AlexNet with only three convolution layers. To do so, we remove the convolution layers C4 and C5 and connect the pooling layer P5 to the output of the convolution layer C3.
We keep the rest of the architecture (early layers and fully connected layers) unchanged.
We notice a drop of accuracy from 71.1\% to 68.6\% for ImageNet-T and 48.7\% to 48.3\% for kth-tips-2b when compared to T-CNN-3.
Therefore our contribution is more than solely using a small network for small datasets as the results show that our method is a good adaptation of CNN to texture analysis.

\mbox{}\\
\begin{table*}[!t]
\caption{Classification results (accuracy \%) for networks trained from scratch. The number of trainable parameters (weights and biases in millions) is indicated in brackets (for 1000 classes).} \label{tab:scratch1}
\centering
\begin{tabular}{|p{2.8cm}|p{1.3cm}|p{1.3cm}|p{1.3cm}|p{1.75cm}|p{1.4cm}|p{1.5cm}|p{1.5cm}|p{1.5cm}|}
\hline
	  & \textbf{Kylberg} & \textbf{CUReT} & \textbf{DTD} &  \textbf{kth-tips-2b} &  \textbf{ImNet-T} &  \textbf{ImNet-S1} &  \textbf{ImNet-S2} &  \textbf{ImageNet} \\ \hline
	 \textbf{T-CNN-1} (20.8) & 89.5 \scriptsize $\pm$1.0 & 97.0 \scriptsize $\pm$1.0 & 20.6 \scriptsize $\pm$1.4 & 45.7 \scriptsize $\pm$1.2  & 42.7 & 34.9 & 42.1 & 13.2 \\ \hline
	 \textbf{T-CNN-2} (22.1) & \textbf{99.2 \scriptsize $\pm$0.3}  & 98.2 \scriptsize $\pm$0.6 & 24.6 \scriptsize $\pm$1.0 & 47.3 \scriptsize $\pm$2.0 & 62.9 & 59.6 & 70.2 & 39.7 \\ \hline
	 \textbf{T-CNN-3} (23.4) & \textbf{99.2 \scriptsize $\pm$0.2} & 98.1 \scriptsize $\pm$1.0 & \textbf{27.8 \scriptsize $\pm$1.2} &  \textbf{48.7 \scriptsize $\pm$1.3} & \textbf{71.1} &  \textbf{69.4} &\textbf{78.6} & 51.2 \\ \hline
	 \textbf{T-CNN-4} (24.7) & 98.8 \scriptsize $\pm$0.2 & 97.8 \scriptsize $\pm$0.9 & 25.4 \scriptsize $\pm$1.3 & 47.2 \scriptsize $\pm$1.4 &  \textbf{71.1} &  \textbf{69.4} & 76.9 &  28.6 \\ \hline
	 \textbf{T-CNN-5} (25.1) & 98.1 \scriptsize $\pm$0.4 & 97.1 \scriptsize $\pm$1.2 & 19.1 \scriptsize $\pm$1.8 & 45.9 \scriptsize $\pm$1.5 & 65.8 & 54.7 & 72.1 & 24.6 \\ \hline
	 \textbf{AlexNet} (60.9) & 98.9 \scriptsize $\pm$0.3 & \textbf{98.7 \scriptsize $\pm$0.6} & 22.7 \scriptsize $\pm$1.3 & 47.6 \scriptsize $\pm$1.4 & 66.3 & 65.7 & 73.1 &  \textbf{57.1} \\ \hline
\end{tabular}
\end{table*}
\begin{table*}[!t]
\caption{Classification results (accuracy \%) using networks pre-trained on ImageNet. The number of trainable parameters (in millions) is indicated in brackets (for 1000 classes).
The state of the art (SoA) results are presented in the last row as reported in the references.} \label{tab:finetune1}
\centering
\begin{tabular}{ |p{2.65cm}|p{2cm}|p{2cm}|p{2cm}|p{2cm}|p{1.4cm}|p{1.5cm}|p{1.5cm}|}
\hline
	  & \textbf{Kylberg} & \textbf{CUReT} & \textbf{DTD} &  \textbf{kth-tips-2b} &  \textbf{ImNet-T} &  \textbf{ImNet-S1} &  \textbf{ImNet-S2} \\ \hline
	\textbf{T-CNN-1} (20.8) & 96.7 \scriptsize $\pm$0.2 & 99.0 \scriptsize $\pm$0.3 & 33.2 \scriptsize $\pm$1.1 & 61.4 \scriptsize $\pm$3.1 & 51.2 & 46.2 & 53.5 \\ \hline
	\textbf{T-CNN-3} (23.4) & \textbf{99.4 \scriptsize $\pm$0.2} & \textbf{99.5 \scriptsize $\pm$0.4} & 55.8 \scriptsize $\pm$0.8 & \textbf{73.2 \scriptsize $\pm$2.2} & 81.2 & 82.1 & 87.8 \\ \hline
	\textbf{AlexNet} (60.9) & \textbf{99.4 \scriptsize $\pm$0.1} & 99.4 \scriptsize $\pm$0.4 & \textbf{56.3 \scriptsize $\pm$1} & 71.5 \scriptsize $\pm$1.4& \textbf{83.2} & \textbf{85.4} & \textbf{90.8} \\ \hline
	\textbf{SoA} & 99.7 \scriptsize \cite{kylberg2013evaluation} & 99.8 \scriptsize $\pm$0.1 \cite{cimpoi2014describing}& 75.5 \scriptsize $\pm$0.8 \cite{cimpoi2015deep} & 81.5 \scriptsize $\pm$2.0 \cite{cimpoi2015deep} & - & - & - \\ \hline
\end{tabular}
\end{table*}
\subsubsection{Network depth analysis}
Tables \ref{tab:scratch1} and \ref{tab:finetune1} show that our approach performs best with three convolution layers (T-CNN-3) with T-CNN-4 and T-CNN-2 close behind.
Averaging the output of convolution layers is like measuring the response to a set of filters.
The more convolution layers we use, the more complex the features sought by these filters are.
In our method, using five layers performs worse as the fifth layer extracts sparse and complex object-like features (nose, wheel etc.).
Averaging the response of these complex features over the entire feature map is not appropriate as such features may be very sparsely found in the input images.
Intuitively one could think that, for this deep architecture, a maximum energy layer would achieve better results than an averaging energy, especially in an object recognition task.
To compare, we test a maximum energy layer which simply outputs the maximum response of each entire feature map. Thus, it measures whether a certain feature was found in the input image,
neither taking into account its location nor the number of occurrences. The results in Table \ref{tab:avemax2} confirm this idea as one can see that shallow T-CNNs perform better with an averaging energy layer while deeper ones are more accurate with a maximum energy layer.

On the other extreme, features in the first layer are too simple (mainly edges) and the network does not have enough learnable parameters to perform as well as T-CNN-3 as can be seen in Tables \ref{tab:scratch1} and \ref{tab:finetune1}.
This depth analysis does not generalize to all network architectures as a very deep approach with a large number of parameters can implement very complex functions and has shown excellent results
in texture classification in \cite{cimpoi2015deep}.
The analysis of such very deep architectures is out of the scope of this paper.
\begin{table}[!t]
\caption{Comparison (accuracy \%) between average and maximum energy layers for various network depths.} \label{tab:avemax2}
\centering
\begin{tabular}{ |p{1.6cm}|p{1.4cm}|p{1.4cm}|p{1.4cm}|p{1.4cm}| }
\hline
	  & \multicolumn{2}{c|}{\textbf{ImageNet-T}} & \multicolumn{2}{c|}{\textbf{ImageNet-S1}} \\ \hline
	 method & average & max & average & max \\ \hline
	\textbf{T-CNN-1} & \textbf{42.7} & 41.3 & \textbf{34.9} & 24.1 \\ \hline
	\textbf{T-CNN-3} & 71.1 & 71.0 & 69.4 & 70.6 \\ \hline
	\textbf{T-CNN-5} & 65.8 & \textbf{67.4} & 54.7 & \textbf{67.6} \\ \hline
\end{tabular}
\end{table}
\subsubsection{Domain transfer}
Table \ref{tab:finetunes3} shows the classification results on kth-tips-2b of networks that were pre-trained on different datasets.
Finetuning a network that was pre-trained on a texture dataset (ImageNet-T) achieves better results on another texture dataset (kth-tips-2b)
as compared to a network pre-trained on an object dataset of the same size (ImageNet-S1 and ImageNet-S2).
T-CNN-3 pre-trained on ImageNet-T results in 61.8\% whereas it drops to 56.3\% and 59.0\% when pre-trained on ImageNet-S1 and ImageNet-S2 respectively.
However, the size of the dataset on which the T-CNN is pre-trained predominantly influences the performance of the finetuning
as one can see on Table \ref{tab:finetunes3}. The network pre-trained on ImageNet significantly outperforms the other pre-trained networks (73.2\% with T-CNN-3).
These two observations suggest that a very large texture dataset could bring a significant contribution to CNNs applied to texture analysis.
\begin{table*}[!t]
\caption{Classification results (accuracy \%) on the kth-tips-2b dataset using networks pre-trained on different databases.} \label{tab:finetunes3}
\centering
\begin{tabular}{ |p{1.6cm}|p{2.2cm}|p{2.2cm}|p{2.2cm}|p{2.2cm}| }
\hline
	 & \textbf{ImageNet} & \textbf{ImageNet-T} & \textbf{ImageNet-S1} & \textbf{ImageNet-S2} \\ \hline
	\textbf{T-CNN-1} & \textbf{61.4 \scriptsize $\pm$3.1} & 54.3 \scriptsize $\pm$2.7 & 52.9 \scriptsize $\pm$2.7 & 53.1 \scriptsize $\pm$3.6 \\ \hline
	\textbf{T-CNN-3} & \textbf{73.2 \scriptsize $\pm$2.2} & 61.8 \scriptsize $\pm$3.5 & 56.3 \scriptsize $\pm$3.9  & 59.0 \scriptsize $\pm$3.0 \\ \hline
	\textbf{AlexNet} & \textbf{71.5 \scriptsize $\pm$1.4} & 56.9 \scriptsize $\pm$3.9 & 55.5 \scriptsize $\pm$3.9 & 58.2 \scriptsize $\pm$2.3 \\ \hline
\end{tabular}
\end{table*}
\subsubsection{Results on larger images}
One advantage of our T-CNN method over a classic CNN such as AlexNet is that the input image sizes must not necessarily be fixed thanks to the energy layer which pools the average response of feature maps of any size.
We can transfer the pre-trained parameters (convolution and fully-connected  layers) and fine-tune our T-CNN on the forest species images of size to 640x640.
Table \ref{tab:large} compares our results with T-CNN-3 to the state of the art. We cannot compare to AlexNet as it requires input images of size 227x227.
Our method outperforms the state of the art \cite{hafemann2014forest} on the Macroscopic forest species dataset (+1.4\%) and obtains very similar results on the Microscopic one (-0.3\%) while using a less complex approach.
\begin{table}[!t]
\caption{Classification results (accuracy \%) on the forest species datasets of the T-CNN-3 and comparison with the State of the Art (SoA).} \label{tab:large}
\centering
\begin{tabular}{ |p{3.3cm}|p{2.2cm}|p{2.2cm}| }
\hline
	 & \textbf{Macroscopic} & \textbf{Microscopic} \\ \hline
	\textbf{SoA} \scriptsize \cite{hafemann2014forest} & 95.77 \scriptsize $\pm$0.27 & \textbf{97.32 \scriptsize $\pm$0.21 } \\ \hline
	\textbf{T-CNN-3} & \textbf{97.2 \scriptsize $\pm$0.40} & 97.0 \scriptsize $\pm$0.61  \\ \hline
\end{tabular}
\end{table}

\subsection{Results combining texture and shape analysis}
Table \ref{tab:combined} shows the improvements obtained by combining the T-CNN to a classic AlexNet network.
Our networks (in bold) are pre-trained on the ImageNet dataset and finetuned on kth-tips-2b.
First, the voting scores of T-CNN-3 and AlexNet (outputs of the softmax layer) are summed to provide an averaged classification of the two networks.
An increase of 0.7\% is obtained in the classification of kth-tips-2b as compared to the T-CNN-3 and 1.6\% as compared to AlexNet.
Finally, the combined network that we name TS-CNN (Texture and Shape CNN) described in part 2.2 obtains the best results with 73.7\%.
It is therefore possible to combine both approaches within the same network to learn texture and overall shape information.
Although our approach does not compete with the state of the art \cite{cimpoi2015deep} due to a large difference in depth, our results are close to this same state of the art using a medium depth CNN (VGG-M).
As shown in Table \ref{tab:combined}, our ``shape'', ``texture'' and ``shape + texture'' approaches obtain similar results. However, our algorithm is fully contained in a CNN architecture and uses a lower number of parameters.
In Table \ref{tab:combined}, we do not indicate the number of trainable parameters of FV/FC-CNN as it differs between pre-training and finetuning and also requires FV encoding as well as training an SVM classifier.
To give an idea for comparison, the VGG-M network contains 101.7 million parameters (for 1000 classes) whereas our T-CNN-3 architecture only has 24 million parameters.

\begin{table*}[!t]
\caption{Classification results (accuracy \%) on kth-tips-2b using AlexNet and T-CNN-3 separately and combined as well as the state of the art method with a medium depth CNN (VGG-M).
The number of trainable parameters (in millions) is indicated in brackets (for 1000 classes).} \label{tab:combined}
\centering
\begin{tabular}{ | p{2cm} | p{6.5cm} | p{1.7cm} |}\hline
\multirow{2}{*}{Shape}  & \textbf{AlexNet} (60.9) & 71.5 \scriptsize $\pm$1.4 \\\cline{2-3}
    & VGG-M FC-CNN \scriptsize \cite{cimpoi2015deep} & 71.0 \scriptsize $\pm$2.3 \\\hline
\multirow{2}{*}{Texture} & \textbf{T-CNN-3} (23.4) & 73.2 \scriptsize $\pm$2.2 \\\cline{2-3}
    & VGG-M FV-CNN \scriptsize \cite{cimpoi2015deep} & 73.3 \scriptsize $\pm$4.7 \\\hline
\multirow{3}{*}{\begin{minipage}{0.5in}Texture and Shape\end{minipage}} & \textbf{sum scores AlexNet T-CNN-3} (84.3) & 73.4 \scriptsize $\pm$1.4 \\\cline{2-3}
    & \textbf{TS-CNN-3} (62.5) & \textbf{74.0 \scriptsize $\pm$1.1} \\\cline{2-3}
    & VGG-M FV+FV-CNN \scriptsize \cite{cimpoi2015deep}  & 73.9 \scriptsize $\pm$4.9 \\\hline
\end{tabular}
\end{table*}

\section{Conclusion}
In this paper we have developed a new CNN architecture for analyzing texture images.
Inspired by classic neural networks and filter banks approaches, we introduce an energy measure which allows us to discard the overall shape information analyzed by classic CNNs.
We showed that with our T-CNN architecture, we can increase the performance in texture recognition, while largely reducing the complexity, memory requirements and computation time.
Finally, we developed a network that incorporates our texture CNN approach into a classic deep neural network architecture, demonstrating their complementarity with a significant improvement of accuracy.

\section{Future research}
Our energy pooling approach enables inputting images of various dimensions, while keeping the exact same structure. Thus, it is possible to implement a multi-scale analysis
by using multiple rescaled images as inputs like in \cite{cimpoi2015deep}. 
Alternatively, the multiscale analysis and computation reduction of the GoogLeNet \cite{szegedy2014going} could be incorporated to our network.

\bibliographystyle{abbrv}
\bibliography{paper_TCNN}
\clearpage

\noindent\textbf{Appendix A: List of classes from ImageNet in ImageNet-T}
\newline
\\n02871525 bookshop, bookstore, bookstall
\\n02927161 butcher shop, meat market
\\n02999410 chain 
\\n03000134 chainlink fence 
\\n03042490 cliff dwelling 
\\n03089624 confectionery, confectionary, candy store 
\\n03207743 dishrag, dishcloth 
\\n03216828 dock, dockage, docking facility 
\\n03461385 grocery store, grocery, food market, market 
\\n03530642 honeycomb 
\\n03598930 jigsaw puzzle
\\n04200800 shoe shop, shoe-shop, shoe store 
\\n04209239 shower curtain 
\\n04326547 stone wall 
\\n04418357 theater curtain, theatre curtain 
\\n04435653 tile roof 
\\n04523525 vault 
\\n04525038 velvet 
\\n04589890 window screen 
\\n04599235 wool, woolen, woollen 
\\n07714571 head cabbage 
\\n07718747 artichoke, globe artichoke 
\\n07831146 carbonara 
\\n09193705 alp 
\\n09332890 lakeside, lakeshore 
\\n09421951 sandbar, sand bar 
\\n11879895 rapeseed 
\\n12144580 corn 

\noindent\textbf{Appendix B: List of classes from ImageNet in ImageNet-S1}
\newline
\\n02085620 Chihuahua 
\\n02099601 golden retriever 
\\n02165456 ladybug, ladybeetle, lady beetle, ladybird, ladybird beetle 
\\n02676566 acoustic guitar 
\\n02701002 ambulance 
\\n02708093 analog clock 
\\n02823750 beer glass 
\\n02860847 bobsled, bobsleigh, bob 
\\n02877765 bottlecap 
\\n02883205 bow tie, bow-tie, bowtie 
\\n02974003 car wheel 
\\n02992529 cellular telephone, cellular phone, cellphone, cell, mobile phone 
\\n03063599 coffee mug 
\\n03124170 cowboy hat, ten-gallon hat 
\\n03187595 dial telephone, dial phone 
\\n03196217 digital clock 
\\n03250847 drumstick 
\\n03255030 dumbbell 
\\n03376595 folding chair 
\\n03388183 fountain pen 
\\n03481172 hammer 
\\n03584254 iPod 
\\n03657121 lens cap, lens cover 
\\n03791053 motor scooter, scooter 
\\n03888605 parallel bars, bars 
\\n03891251 park bench 
\\n03970156 plunger, plumber's helper 
\\n04356056 sunglasses, dark glasses, shades 
\newline

\noindent\textbf{Appendix C: List of classes from ImageNet in ImageNet-S2}
\newline
\\n01677366 common iguana, iguana, Iguana iguana
\\n01807496 partridge
\\n02099267 flat-coated retriever
\\n02108422 bull mastiff
\\n02113712 miniature poodle
\\n02130308 cheetah, chetah, Acinonyx jubatus
\\n02319095 sea urchin
\\n02480495 orangutan, orang, orangutang, Pongo pygmaeus
\\n02483362 gibbon, Hylobates lar
\\n02786058 Band Aid
\\n02807133 bathing cap, swimming cap
\\n03000684 chain saw, chainsaw
\\n03095699 container ship, containership, container vessel
\\n03532672 hook, claw
\\n03544143 hourglass
\\n03599486 jinrikisha, ricksha, rickshaw
\\n03649909 lawn mower, mower
\\n03710637 maillot
\\n03761084 microwave, microwave oven
\\n03803284 muzzle
\\n03832673 notebook, notebook computer
\\n03840681 ocarina, sweet potato
\\n03854065 organ, pipe organ
\\n03887697 paper towel
\\n04136333 sarong
\\n04501370 turnstile
\\n06785654 crossword puzzle, crossword
\\n12057211 yellow lady's slipper, yellow lady-slipper, Cypripedium calceolus, Cypripedium parviflorum
\end{document}